\documentclass[conference]{IEEEtran}
\IEEEoverridecommandlockouts
\usepackage{cite}
\usepackage{amsmath,amssymb,amsfonts}
\usepackage{algorithmic}
\usepackage{graphicx}
\usepackage{booktabs}
\usepackage{textcomp}
\usepackage{xcolor}
\usepackage{booktabs}
\usepackage{array}
\usepackage[colorlinks=true, linkcolor=black, citecolor=black, urlcolor=black]{hyperref}
\def\BibTeX{{\rm B\kern-.05em{\sc i\kern-.025em b}\kern-.08em
    T\kern-.1667em\lower.7ex\hbox{E}\kern-.125emX}}
\begin{document}

\title{End-to-End Text-to-SQL with Dataset Selection: Leveraging LLMs for Adaptive Query Generation}

\author{
\IEEEauthorblockN{
Anurag Tripathi\IEEEauthorrefmark{1},
Vaibhav Patle\IEEEauthorrefmark{1},
Abhinav Jain\IEEEauthorrefmark{1},
Ayush Pundir\IEEEauthorrefmark{1},
Sairam Menon\IEEEauthorrefmark{2},
Ajeet Kumar Singh\IEEEauthorrefmark{1}\\
Dorien Herremans\IEEEauthorrefmark{3}
}

\IEEEauthorblockA{\IEEEauthorrefmark{1}Infoorigin Pvt Ltd, Data Science\\
Email: \{anurag.tripathi, vaibhav.patle, abhinav.jain, ayush.pundir, ajeetkumar.singh \}@infoorigin.com}

\IEEEauthorblockA{\IEEEauthorrefmark{2}J\&J Innovative Medicine Technology R\&D, USA\\
Email: smenon18@its.jnj.com}

\IEEEauthorblockA{\IEEEauthorrefmark{3}Singapore University of Technology and Design, Singapore\\
Email: dorien\_herremans@sutd.edu.sg}
}


\maketitle 
\begin{abstract}

Text-to-SQL bridges the gap between natural language and structured database language, thus allowing non-technical users to easily query databases. Traditional approaches model text-to-SQL as a direct translation task, where a given Natural Language Query (NLQ) is mapped to an SQL command. Recent advances in large language models (LLMs) have significantly improved translation accuracy, however, these methods all require that the target database is pre-specified. This becomes problematic in scenarios with multiple extensive databases, where identifying the correct database becomes a crucial yet overlooked step.
In this paper, we propose a three-stage end-to-end text-to-SQL framework to identify the user's intended database before generating SQL queries. Our approach leverages LLMs and prompt engineering to extract implicit information from natural language queries (NLQs) in the form of a ruleset. We then train a large db\_id prediction model, which includes a RoBERTa-based finetuned encoder, to predict the correct Database identifier (db\_id) based on both the NLQ and the LLM-generated rules. Finally, we refine the generated SQL by using critic agents to correct errors. Experimental results demonstrate that our framework outperforms the current state-of-the-art models in both database intent prediction and SQL generation accuracy.

\end{abstract}

\begin{IEEEkeywords}
RoBERTa : Robustly Optimized BERT Approach, LLM: Large language Model, SQL: Structured  Query Language, text-to-SQL
\end{IEEEkeywords}

\section{Introduction}
Recent advancements in large language models (LLMs) have unlocked numerous opportunities across various domains, including summarization, code generation, and more. One such opportunity is text-to-SQL, which enables the conversion of natural language queries (NLQs) into SQL, facilitating seamless interaction with structured databases.
State-of-the-art approaches treat text-to-SQL as a machine translation problem, where an NLQ is translated into a corresponding SQL query. This provides a natural language interface for database interaction, ultimately eliminating the need for a human expert in routine database-related tasks. 
The research trajectory of text-to-SQL has evolved significantly over time. Initially, rule-based approaches dominated the field, followed by domain-specific models leveraging sequence-to-sequence deep learning architectures. With the recent advancements in large language models (LLMs), text-to-SQL models have now surpassed all previous approaches, achieving state-of-the-art performance in translating NLQ into SQL. These state-of-the-art (SOTA) text-to-SQL models leveraging large language models (LLMs) \cite{b1} \cite{b2} \cite{b3} have introduced a stage-wise decomposed pipeline. This decomposition has given rise to a new paradigm where each sub-task is responsible for a specific function, including pre-processing, SQL generation, and post-processing. The post-processing stage often involves techniques like self-correction \cite{b2} and self-debugging \cite{b4} \cite{b5} to refine and correct the generated SQL, ensuring higher accuracy and robustness. 
Most LLM-based text-to-SQL approaches currently require the Database ID $(db\_id)$ to be provided manually for schema linking along with the NLQ. However, this approach is not scalable when dealing with large and diverse databases across multiple domains with varying structures. This limitation highlights a significant area for improvement, particularly in the pre-processing and post-processing stages, where automated database identification and schema understanding could enhance the overall efficiency and applicability of text-to-SQL systems.
The main contributions are as follows:
\begin{itemize}
\item End-to-end framework for text-to-SQL that includes dataset prediction. 
    \item An elaborated prompt-based rules generation module to capture implicit information from training data. These rules set the implicit values for possible database Entities to \textit{True} if it presents otherwise \textit{False}
    \item A RoBERTa-based hybrid db\_id prediction model trained with NLQ and implicit information generated by the LLM.
    \item An engineered prompt for SQL generation as well as SQL correction rules for post-processing.
\end{itemize}

\section{Preliminaries}
\subsection{Problem Definition}
Given a natural language question $q$ and a relational database $D$ with a db\_id, the database schema of $D$, denoted as $S$, consists of multiple components. The set of tables in the schema is represented as 
\begin{equation} T = \{T_1, T_2, \dots, T_n\},
\end{equation}
where each $T_i$ corresponds to a table in the database. Each table $T_i$ contains a set of columns denoted as  
\begin{equation}
    C = \{ c^1_1, \dots, c^1_{m_1}, c^2_1, \dots, c^2_{m_2}, \dots, c^n_1, \dots, c^n_{m_n} \},
\end{equation}
where $c^i_j$ represents the $j$-th column of table $T_i$, and $m_i$ denotes the number of columns in $T_i$. Additionally, the schema includes a set of foreign key relationships.
\begin{equation}
    R = \{ (c^i_j, c^k_h) \mid c^i_j, c^k_h \in C \},
\end{equation}
where each pair $(c^i_j, c^k_h)$ defines a foreign key relation between table $T_i$ and table $T_k$.The text-to-SQL task involves learning a mapping function $f$, which takes the natural language query $q$ and the schema $S$ as input and generates a corresponding SQL query that can be executed on $D$. This mapping can be formally expressed as:
\begin{equation}
    f: (q, S) \mapsto \text{SQL},
\end{equation}
where $f$ ensures that the SQL query correctly retrieves the desired information from the relational database.

\subsection{Framework} The proposed end-to-end text-to-SQL framework with Data Selection, as shown in Fig.~\ref{fig:train_pipeleinel}, can be viewed as three subtasks: 
1) Entity generation for db\_id prediction and schema linking using db\_id 2) SQL generation using NLQ and schema for the given NLQ 3) Post-processing of generated SQL using self-correction rules.

We use db\_id prediction rules to generate the entities e.g. gas\_station\_operations, etc., as shown in \ref{fig:cls_prompt} and set True or False based on the presence or absence of entities in the NLQ. Now these entities are represented as a one-hot vector and concatenated with the input NLQ as shown in Fig.~\ref{fig:nw_arch}. This concatenated vector $q'$ serves as input to the db\_id prediction model. Using the predicted db\_id we link the appropriate database schema and a text-to-SQL prompt is constructed on-the-fly to generate SQL for a given NLQ. As a post-processing step, we pass the generated SQL through the multi-agent self-correction module, as shown in Fig.~\ref{fig:SQL_correction}, which finally generates an SQL command for the downstream task.

\section{Related work}

The goal of text-to-SQL is to automatically convert queries in natural language into SQL queries. In addition to significantly increasing data processing efficiency and bridging the gap between non-expert users and database systems, it supports a broader range of applications, including intelligent database services, automatic data analysis, and database question-answering. However, because it might be difficult to properly comprehend NLQ and generate accurate SQL queries, text-to-SQL remains a challenging task \cite{b7}\cite{b8}. Numerous studies on text-to-SQL have been carried out in the fields of natural language processing and databases. Early research either treated the text-to-SQL problem as a sequence-to-sequence task with an encoder-decoder architecture \cite{b9}\cite{b10} or approached it using pre-defined rules or query enumeration \cite{b11}\cite{b12}. Many methods, including attention mechanisms \cite{b13}, graph representation \cite{b14}\cite{b15} syntax parsing \cite{b16}\cite{b3} etc., are used to aid text-to-SQL tasks as deep learning advances quickly. We used one of the widely popular Deep learning pre-trained RoBERTa-based models for the db\_id Prediction  task\cite{b17}\cite{b15}. Additionally, several extensive benchmark datasets have been made publicly available to bridge the gap between text-to-SQL research and its practical implementation. In this paper, we use spider data for our experiment \cite{b18}. 
\subsection{LLMs for text-to-SQL}
For many years, the natural language processing (NLP) and database communities have been actively researching text-to-SQL, or the conversion of NLQ into SQL queries \cite{b19}\cite{b17}. Large Language Models' (LLMs') potential efficacy has recently helped text-to-SQL \cite{b20} For SQL generation, early techniques made use of LLMs' zero-shot in-context learning capabilities \cite{b21}. Using task decomposition and methods like Chain-of-Thought (CoT) \cite{b22}, later models have enhanced LLM performance. These models include but are not limited to, DAILSQL~\cite{b1}, MAC-SQL \cite{b23}, C3 \cite{b24}, self-consistency \cite{b23}, and least-to-most prompting. Furthermore, research is being done on optimizing LLMs for text-to-SQL \cite{b1,b3}. Based on these works we choose MAGIC~\cite{b28} as the self-correction method for failed cases.

\subsection{Agents Based on LLM: }

Numerous studies have been conducted on autonomous agents based on LLMs, including AutoGPT (Significant Gravitas), OpenAgents \cite{b25}, and AutoGen \cite{b26}, as LLM-based agents have long been a promising field of study in both academic and industry communities~\cite{b27}. There are, however, few works that use this idea for text-to-SQL, and the sole multi-agent approach that focuses on solving the text-to-SQL challenge utilizing a novel multiagent collaborative framework is MAC-SQL\cite{b23}, similarly in \cite{b28} suggests an innovative agent technique that iterates over failed queries in the text-to-SQL process to automatically provide self-correcting guidance. By offering data-driven, 
Inspired by research on automated solutions that replicate human experts' rectification processes, our work integrates a multi-agent framework within the pipeline. This approach enables systematic error correction and refinement, enhancing the overall robustness and accuracy of the generated SQL queries.
To update incorrect predicted SQL, we leverage MAGIC's~\cite{b28} multi-agent framework, which consists of a feedback and correction agent, to iteratively refine the erroneous query using both the predicted and ground truth SQL. This automated self-correction approach reduces reliance on manually crafted guidelines, leading to more effective error mitigation and improved SQL generation performance.

\section{Methodology}
In this section, we describe the proposed framework in detail. We have decomposed the proposed end-to-end text-to-SQL Framework (Fig.~\ref{fig:train_pipeleinel}) into three parts: training db\_id prediction steps, text-to-SQL generation (NLQ to SQL) and SQL self-correction module. Moreover, we divided it into the following sub-tasks.

\begin{itemize}
    \item Training data Generation
    \item Text-to-SQL Pipeline
    \begin{itemize}
        \item db\_id prediction rules Generation
        \item db\_id prediction model Training
        \item Prompt-based SQL Generation
        \item SQL Self-Correction Module
    \end{itemize}
    \item Inference pipeline
\end{itemize}

\subsection{Training Data Preparation}\label{sect:train data prep} The Spider dataset\cite{b18} has train, test and validation  splits with non-overlapping db\_id. This makes it difficult to properly evaluate our method, as it would be trained in an unsupervised setting. Hence, we decided to create a new training, test and validation sets with a split of 70\%, 15\%, 15\% with an equal distribution of db\_ids.
 this new split is essential, otherwise a db\_id Prediction model trained solely on the training set would remain unaware of the classes present in the validation and test sets, thereby limiting its generalization capability. 
\par {Initial Dataset Overview} 
: The dataset initially consisted of 140 distinct classes. However, certain classes contained a limited number of NLQs, making them less robust for db\_id Prediction Model. To address this, a merging strategy was implemented based on predefined rules.
\par {Rule-Based Merging of Classes}:
To ensure a more balanced dataset, classes were merged based on logical groupings. For example: 
\begin{itemize}
    \item {Ex 1.} college\_1, college\_2, college\_3 were merged into a new class named college. 
    \item {Ex 2.} department\_management, department\_store were merged into department
\end{itemize}
\par {Semantic Similarity-Based Merging}
Further optimization was performed by analyzing the semantic similarity of classes using pre-trained sentence transformer (used all-MiniLM-L6-v2 model). This approach made it possible to identify closely related classes that could be merged. Based on this analysis, the following changes were made: 
\begin{itemize}
    \item musical, music, theme were merged into a single class named music\_musical\_theme. 
    \item university, school, college, activity were merged into university\_school\_college\_activity. 
\end{itemize}
Finally, by strategically merging classes through rule-based and embedding-based methods, we achieved a refined dataset with 86 well-defined classes, ensuring improved model performance and generalization. This structured approach helped in creating a more balanced and meaningful db\_id prediction system, optimizing the dataset for further training tasks.

\subsection{Text-to-SQL Pipeline}

\subsubsection{\textbf{db\_id Prediction Rules Generation using Prompt}}
\begin{figure}[ht]
\centerline{\includegraphics[width=\columnwidth]{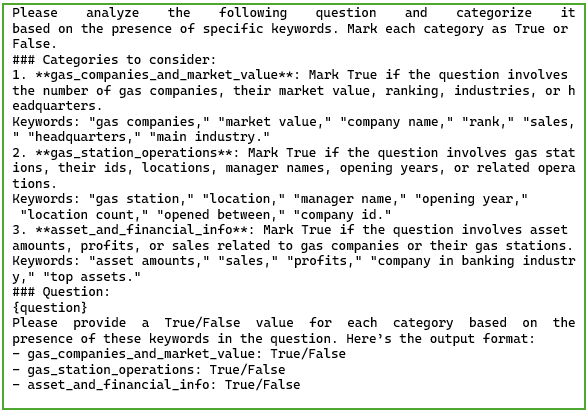}}
\caption{db\_id Prediction Rules Generation Prompt}
\label{fig:cls_prompt}
\end{figure}

For each db\_id in the dataset, a set of rules is defined to predict the  db\_id of the question. The example Fig.~\ref{fig:cls_prompt} illustrates the rule structure for the Gas class. In step 1 and step 2 we are generating the db\_id prediction rule.

step 1. {Input:} NLQ $q$ and Database $D$ with Schema $T$
    
step 2. {Generate db\_id prediction Rules:} 
$R = f_{\text{LLM}}(q)$
\[
R = \begin{bmatrix} R_1,R_2, & \dots & R_k \end{bmatrix}^T = f_{\text{LLM}}(q), \quad R \in \mathbb{R}^k  dimension
\]

\subsubsection{\textbf{db\_id Prediction Model} }
This db\_id prediction process is applied to all questions in the dataset. Each question is processed through these rule-based prompts, producing a set of language entities (True/False) for the predefined rules shown in Steps 3 to 6 in Equation 5. These labeled outputs are then used to generate updated versions of the questions, incorporating the relevant db\_id Prediction into a one-hot encoded format. 

The \textbf{db\_id prediction entity generation workflow} is shown below-

Step 3. {Identify Possible Entities:} $E = g_{\text{LLM}}(q, R)$

         Using another prompt, the LLM determines a set of possible entities to which the NLQ could belong.
         
     where $ {g_{\text{LLM}}} $ maps the query and rules to a set of potential entities.
    
Step 4. {Determine True Entity:} $e^* \in E$

Step 5. {Update NLQ with True Entity:}
\begin{equation}\label{updatedNLQ}
    q' = q \oplus e^*
\end{equation}

Step 6. {Train db\_id Prediction model :} 
\[
        \min_{\theta} \sum_{(q', e^*) \in D} \mathcal{L}(h_{\text{RoBERTa}}(q'; \theta))
 \]
where $\mathcal{L}$ is the loss function and $\theta$ represents the model parameters.

The following demonstrates an example of db\_id language entities mapping and model training.
\begin{itemize}
    \item Input Question: ``Show all locations with only 1 station." 
    \item Processed Results: 
    \begin{itemize}
        \item gas\_companies\_and\_market\_value: True   
        \item gas\_station\_operations: True   
        \item asset\_and\_financial\_info: False 
    \end{itemize}

\item Updated Question for model training: Show all locations with only 1 station. gas\_companies\_and\_market\_value, gas\_station\_operations. 
The updated questions and their corresponding db\_id are then used as training data for the db\_id prediction model. 

\end{itemize}

\begin{figure}[ht]
\centerline{\includegraphics[width=0.50\columnwidth]{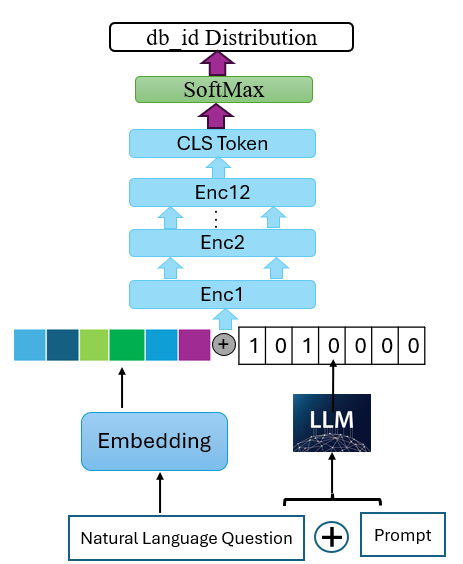}}
\caption{Encoder Architecture for db\_id Prediction Model Training }
\label{fig:nw_arch}
\end{figure}

\begin{figure}[ht]
\centerline{\includegraphics[width=\columnwidth]{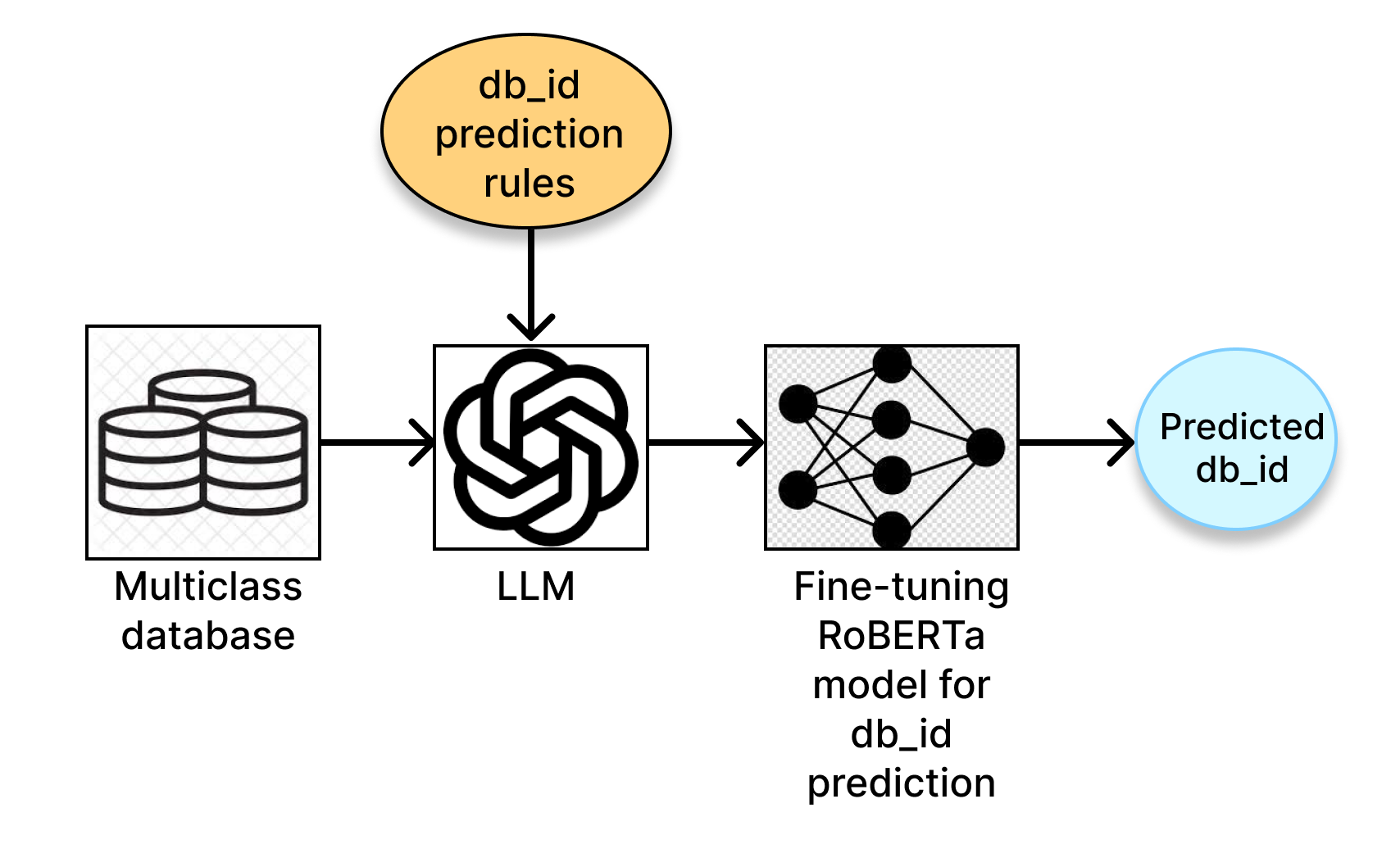}}
\caption{Encoder based db\_id Prediction Model Training Pipeline}
\label{fig:cls_model}
\end{figure}

\textbf{Finetuning the db\_id Prediction Model:} 
For our db\_id Prediction task, we utilized a transformer-based approach with RoBERTa as an encoder-only model. RoBERTa (Robustly Optimized BERT Pretraining Approach) is a variant of BERT that improves contextual representations through optimized pre-training strategies such as dynamic masking, increased training data, and larger batch sizes. Designed to enhance performance through these optimizations, RoBERTa's contextualized representations help improve the accuracy and robustness of our db\_id prediction model. 
The output of the db\_id Prediction process (updated questions + db\_id) are used to tune the db\_id Prediction Model, as shown in Fig.~\ref{fig:nw_arch}. The model learns to predict the appropriate db\_id for a given question based on its db\_id Prediction-enhanced format. 
The db\_id  Prediction Model Training Pipeline is shown in Fig.~\ref{fig:cls_model}. Once trained, the db\_id Prediction task model becomes an integral part of the end-to-end pipeline, enabling accurate db\_id prediction during runtime.

\subsubsection{\textbf{Prompt based SQL generation}}
\begin{figure}[ht]
\centerline{\includegraphics[width=\columnwidth]{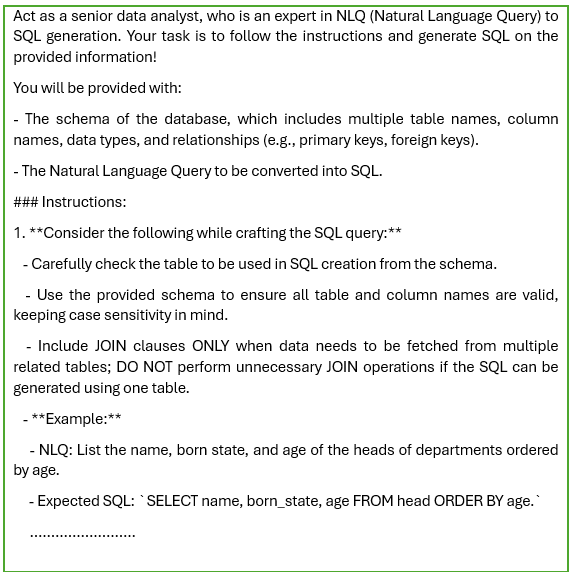}}
\caption{SQL Generation Prompt}
\label{fig:SQL_promt}
\end{figure}

This section outlines the implementation of an end-to-end pipeline for converting NLQ into SQL queries using advanced language models and a db\_id Prediction Model. The pipeline ensures schema linking identification and SQL query generation by leveraging text-to-SQL prompts. shown in Fig.~\ref{fig:SQL_promt} and db schema. After predicting the db\_id, the pipeline proceeds to SQL generation, as shown in steps 1 to 5. The steps are as follows:-

{Step 1: Input Natural Language Query:-}
Let $q$ be the given natural language query:
\begin{equation}
    q \in \mathcal{Q}
\end{equation}
where $\mathcal{Q}$ represents the space of all NLQs.

{Step 2: Append True Entity:-}
Each NLQ is associated with a true entity $e^*$, which is appended to form an updated query:
\begin{equation}
    q' = q \oplus e^*
\end{equation}
where $\oplus$ represents string concatenation.

{Step 3: Classify the NLQ:-}
db\_id Prediction model $h_{\text{cls}}$ maps the query $q'$ to a class label $c$:
\begin{equation}
    c = h_{\text{cls}}(q')
\end{equation}
where $c \in \mathcal{C}$ and $\mathcal{C}$ is the set of possible query classes.

\begin{itemize} 
     \item {Schema Extraction:} The schema corresponding to the predicted db\_id is retrieved. 

     {Step 4: Schema Selection-}
A function $\sigma$ maps the query class $c$ to a database schema $s^*$:
\begin{equation}
    s^* = \sigma(c)
\end{equation}
where $S = \{s_1, s_2, \dots, s_M\}$ is the set of available schemas, and $s^* \in S$ is the most relevant schema.

    \item {Prompting for SQL Generation:} The updated question, schema, and predefined SQL generation prompt are passed to the language model. The model generates the SQL query based on these inputs.

{Step 5: SQL Query Generation}
A large language model (LLM), $f_{\text{LLM}}$, generates the SQL query using $q'$ and the selected schema $s^*$:
\begin{equation}
    \text{SQL} = f_{\text{LLM}}(q', s^*)
\end{equation}
where $\text{SQL} \in \mathcal{SQL}$, the space of valid SQL queries.
The complete text-to-SQL transformation can be expressed as:
\begin{equation}
    \text{SQL} = f_{\text{LLM}}(q' , \sigma(h_{\text{cls}}(q')))
\end{equation}
where:
     $q' = q \oplus e^*$ (query augmentation),
     $h_{\text{cls}}(q')$ classifies the query,
      $\sigma(h_{\text{cls}}(q'))$ selects the appropriate schema,
      $f_{\text{LLM}}$ generates the final SQL query.
      
    \item {Accuracy Evaluation:} The generated SQL query is compared with the ground truth SQL query to evaluate the pipeline's accuracy.
\end{itemize}
\begin{figure*}[ht]
\centerline{\includegraphics[width=1.96\columnwidth]{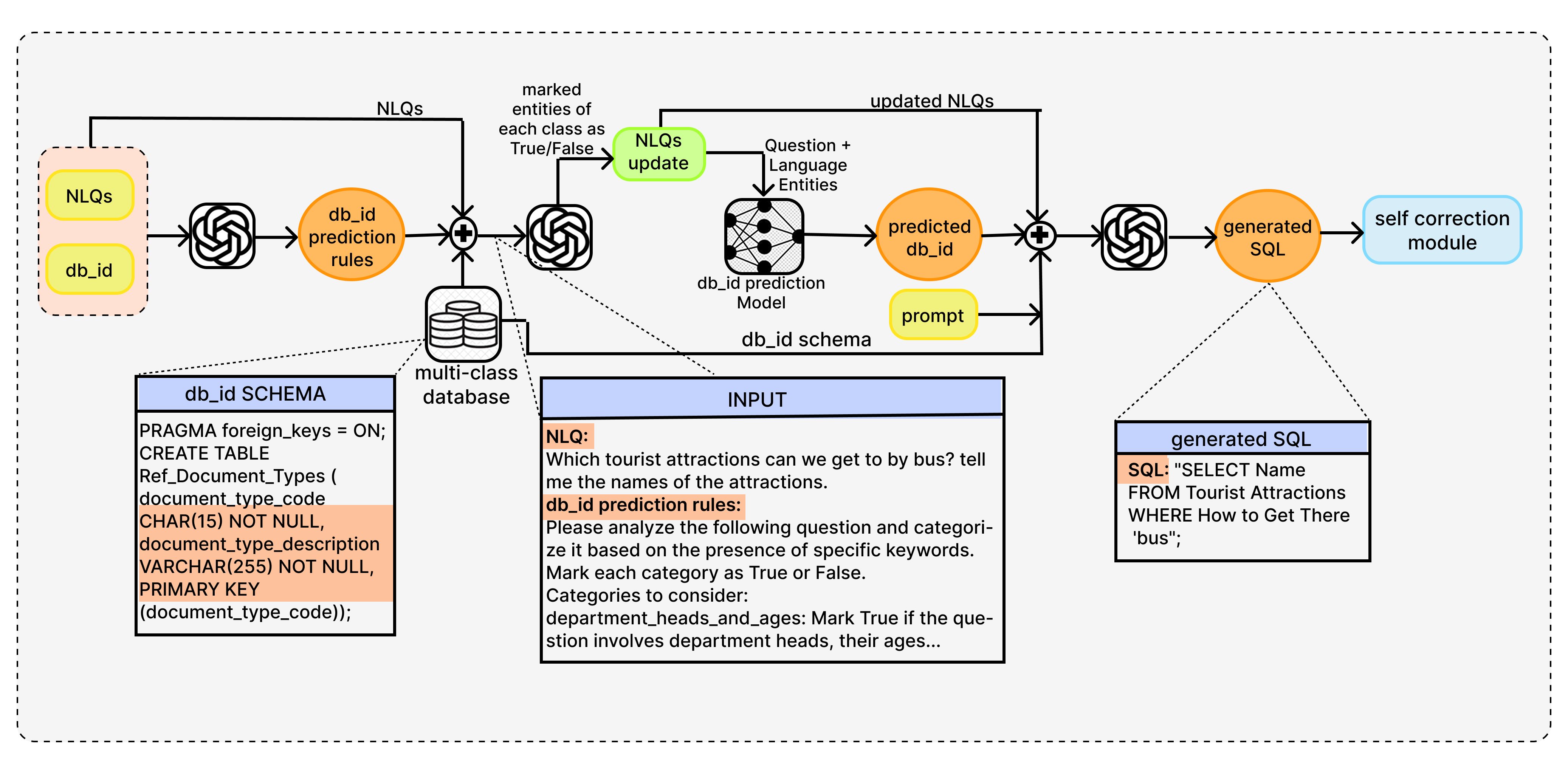}}
\caption{\textbf{End-to-End text-to-sql Framework} 
    The Framework consists of three components: db\_id prediction rules generation, db\_id prediction model training and SQL Self-Correction Module. The initial two components involves creating rules that map NLQs to db\_id and training model with above rules which could be used at inference, while the latter focuses on generating SQL queries that allow the self-correction module to refine its output, correcting any misinterpretations or errors in the SQL query generated.}
\label{fig:train_pipeleinel}
\end{figure*}

\subsubsection{\textbf{SQL Self-Correction Module with Feedback and Correction Agents}} \label{sect:SQL-self-correction} 
\begin{figure}[ht]
\centerline{\includegraphics[width=1.0\columnwidth]{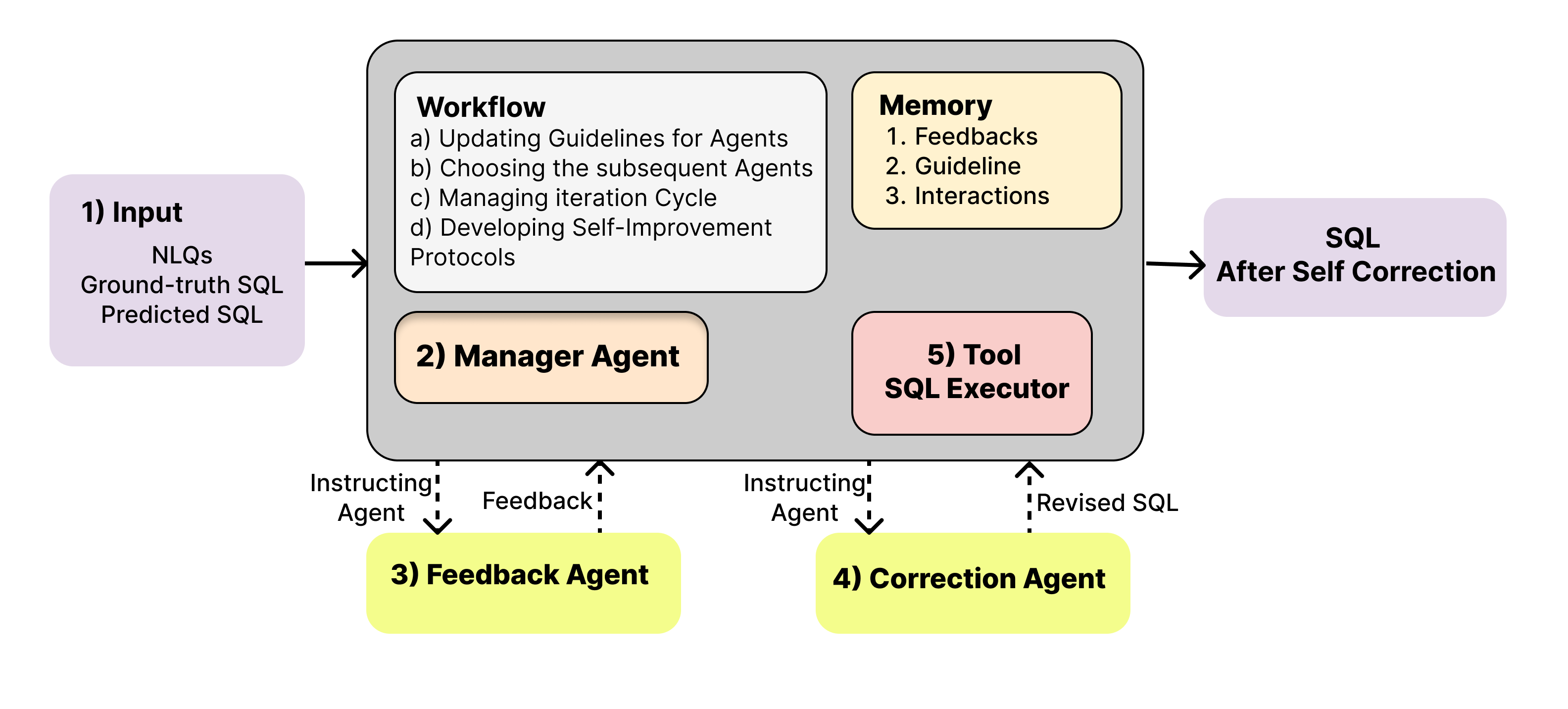}}
\caption{\textbf{SQL Self-Correction Module} The module comprises three key components: a Feedback Agent,
Correction Agent and Manager Agent designed to refine and self-correct the predicted SQL.}
\label{fig:SQL_correction}
\end{figure}

The SQL self-generation module, as shown in Fig.~\ref{fig:SQL_correction}, comprises three key components: a Feedback Agent, Correction Agent and Manager Agent designed to refine SQL query generation iteratively. The Feedback Agent takes as input the correct ground truth SQL and a corresponding incorrectly predicted SQL query, systematically analyzing discrepancies between them to identify error patterns. These identified issues are then passed to the Correction Agent, which not only updates the predicted SQL query to align with the ground truth but also formulates structured correction guidelines. The Manager Agent is responsible for choosing subsequent Agents, managing the iteration cycle and updating the Guidelines for SQL Self-Correction. These guidelines serve as a knowledge base that informs future inference-time SQL generation, enabling the model to self-improve by leveraging past correction insights. This approach ensures robustness in SQL generation by integrating a continual feedback loop that enhances query accuracy over successive iterations.

\subsection{Inference Pipeline} 
The inference pipeline is designed to tranform NLQ into accurate SQL statements through a multi-step process. Initially, the \textbf{db\_id Prediction model}, as shown in Fig.\ref{fig:cls_model}, and rules, which were developed during the training phase, is utilized to map components of the user's query to relevant database schema elements, such as table names, column names, and data types. These rules were learned during the training phase, enabling the model to understand the structure of the database and effectively align natural language expressions with corresponding database identifiers. During inference, the pipeline simply applies these pre-established rules to ensure that the db\_id is predicted correctly reducing ambiguity and potential mismatches between the user's intent and the database identifier.

Following, when generating the SQL from the text prompt, the \textbf{SQL Self-Correction module}, as shown in Fig.\ref{fig:SQL_correction}, is employed to correct SQL queries. Similarly to the db\_id Prediction Rules, the mechanisms for identifying and correcting SQL errors were established during the training phase. At inference time, this module leverages those learned strategies to detect and adjust any inaccuracies or misinterpretations in the initial SQL output.

\section{Experimental setup}
In this work, we have proposed an end-to-end text-to-SQL Framework in Fig. ~\ref{fig:train_pipeleinel} which can be decomposed into a db\_id Prediction Model, SQL Generation and SQL Self-Correction. As compared to the state-of-the-art, where schema linking happens through manually associated db\_id,  our Framework predicts the db\_id using a db\_id Prediction Model followed by Prompt-based SQL generation and SQL Self-Correction. 

\subsection{Models}
We have evaluated our proposed framework using a hybrid RoBERTa-base db\_id Prediction Model, as shown in Fig.~\ref{fig:nw_arch}. For prompt-based rule learning and text-to-SQL generation we have used two variants GPT family (GPT-4o-mini and GPT-3.5-turbo). 
\subsection{Hyper-parameters} The Adam optimizer with a learning rate of \(1.0 \times 10^{-5}\) was used with cross-entropy loss.
The model was trained with a batch size of 8. Additionally, we leveraged GPT-4o-mini and GPT-3.5-turbo via the OpenAI API for generating the db\_id Prediction rules, SQL correction rules, and the final SQL generation. This allowed us to enhance performance and perform a comparative analysis.
\subsection{Dataset}
The Spider dataset comprises three distinct subsets: development \(dev\),  \(test\) and  \(train\). 
	
It is important to note that the db\_ids in the training, test, and validation sets are mutually exclusive. To ensure that our db\_id Prediction model is exposed to all possible classes, we have merged these three subsets. The final test set constitutes 1,058 NLQs as shown in Table~\ref{tab:nlSQL_count}.
In an initial test, we found that the NLQs for which our solution could not provide predicted SQL were often controversial or missing important information. There were certain NLQs for which the db\_ids  schema file is missing or it contains an incorrect schema file name. Hence, we performed postprocessing and removed these from the dataset, to obtain our final test set.

\subsection{Metrics} We use Mean Average Precision (MAP), NDCG  (Normalized Discounted Cumulative Gain), Precision and Recall to gauge db\_id Prediction accuracy of the db\_id prediction model. To evaluate the text-to-SQL model, we use  execution accuracy (EX) and exact set match (EM). 

\textbf {Mean Average Precision (MAP)} \cite{b6} is useful in multi-label db\_id Prediction where multiple labels can be correct for each instance.

\textbf{NDCG} (Normalized Discounted Cumulative Gain) is another evaluation metric used in ranking problems. Unlike MAP, which considers binary relevance, NDCG accounts for graded relevance (e.g., highly relevant, somewhat relevant, not relevant).

\textbf{Precision@1} (P@1) is a ranking metric that evaluates how often the top-1 predicted label is correct. It is commonly used in multi-label db\_id Prediction and ranking tasks to measure how well a model ranks the most relevant item first

\textbf{Recall@1} (R@1) measures how often the top-1 predicted label is correct out of all the actual relevant labels. It evaluates the ability of the model to retrieve at least one correct label when only the highest-ranked prediction is considered.

\textbf{Exact Set Match (EM)} \cite{b18} checks if the generated SQL query matches the reference query exactly in terms of syntax and structure. It is a stricter metric compared to Execution Accuracy because even minor differences in the query (like formatting or aliasing) can lead to failure in exact match, even if the query executes correctly.

\textbf{Execution Accuracy (EX)} \cite{b18} measures whether the SQL query, when executed on the database, returns the correct result. Even if the query structure differs from the reference query, as long as the result is correct, it is considered accurate.

\section{Results}
The results of our study are three-fold, encompassing both db\_id prediction model outcomes, text-to-SQL performance, and SQL self-correction module. The db\_id prediction model demonstrates robust accuracy on metrics as detailed in Table \ref{tab:dbid accuracy}, highlighting its capability to effectively identify relevant database schemas. For the text-to-SQL component, we evaluate performance using Execution Accuracy (EX) and Exact Match (EM) metrics as detailed in Table \ref{tab:SQL_Accuracy}.

\subsection{db\_id Prediction Results}
The Table  \ref{tab:dbid accuracy} presents the evaluation metrics for the db\_id Prediction Model shown in Fig.~\ref{fig:cls_model} and DeBERTa model\cite{b6} in predicting db\_ids based on NLQ. The model was tested on 1,058 NLQ instances, out of which 1,036 predictions fell within the top 5 ranked results, while 21 were beyond the top 5. The model demonstrated strong performance with an NDCG (Normalized Discounted Cumulative Gain) of 0.945 and a Mean Average Precision (MAP) of 0.933, indicating high-ranking quality and retrieval effectiveness. Furthermore, Our Pipeline also showed strong precision and recall at rank 1, with Precision@1 of 0.901 and Recall@1 of 0.899, but these values were not reported for DeBERTa V2 xlarge\cite{b6}.\par
Overall, both models demonstrated high retrieval performance, but our  db\_id Prediction model was tested on a significantly larger dataset, proving its robustness in handling a more extensive range of queries.

\begin{table}[ht] 
\caption{NLQ-SQL data count} 
\label{tab:nlSQL_count}
\centering 
\begin{tabular}{@{}lc@{}} 
\toprule 
\textbf{Quantity }                                &\textbf{Count}  \\ \midrule 
Initial NLQs in Test Set                 & 1058  \\  
NLQs after removing Missing Schema File  & 965   \\ 
After removing SQLs not executed         & 781   \\ 
SQLs executed with correct output        & 722   \\ 
\bottomrule 

\end{tabular} 
\end{table}

\begin{table*}[htbp!] 

\caption{\centering db\_id Prediction accuracy metric}
\label{tab:dbid accuracy}
\footnotesize
\begin{center}
\begin{tabular}{lccccccc}
\toprule
\textbf{MODEL}   &\textbf{Test NLQ} & \textbf{Within Top 5} &\textbf{ More than top5} & \textbf{NDCG} & \textbf{MAP} & \textbf{Precision @1} &\textbf{Recall @1} \\ \midrule
\textbf{db\_id Prediction Model (our)} & 1058 & 1036 & 21 & 0.945 & 0.933 & 0.901 & 0.899 \\ 
DeBERTa V2 xlarge & 159 & 144 & 15 & 0.947 & 0.938 & - & - \\ \bottomrule
\end{tabular}

\end{center}
\end{table*}

\subsection{SQL Generation Results}
We evaluated our proposed framework on the Spider dataset \cite{b18} and compared its performance against SOTA approaches using exact-set-match (EM) and execution accuracy (EX) metrics, as presented in Table \ref{tab:SQL_Accuracy}. This table highlights the performance of various methods, including prompt engineering and in-context learning, providing a comprehensive comparison of text-to-SQL approaches.
Since the same SQL can have multiple syntactical representations, Hence, EX is the most relevant metric to evaluate Text-to-SQL Model as compared to EM.
Although EM is not very relevant to quantify a framework, still we have shown in \ref{tab:SQL_Accuracy} that our method has a comparable accuracy to other SOTA methods and we obtain superior results within the class of Prompt Engineering-based approaches\cite{b1,b2,b31}

\begin{figure}[ht]

\centerline{\includegraphics[width=\columnwidth]{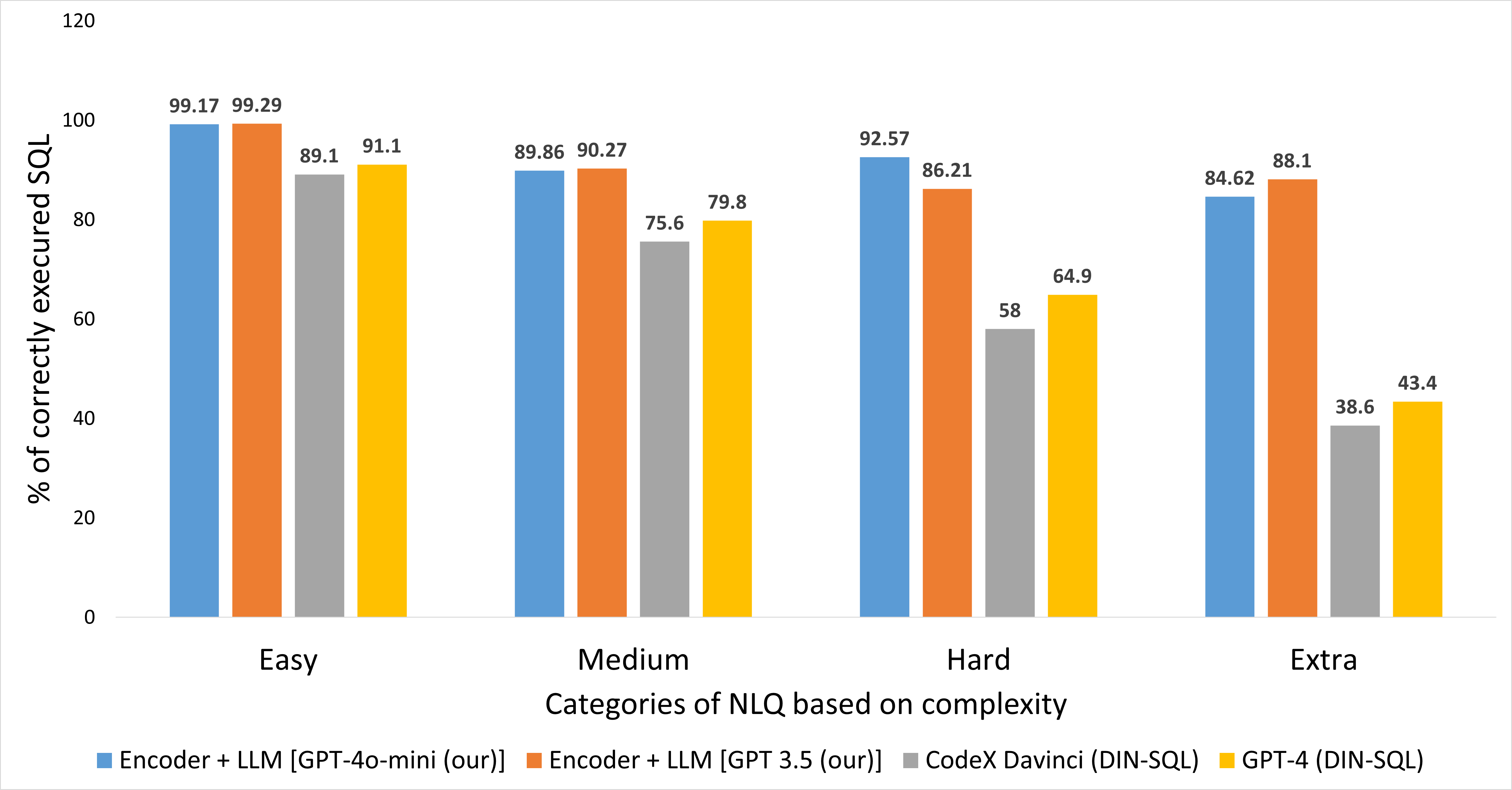}}
\caption{\textbf{Category wise SQL generation results:} The above chart depicts that the results are categorized by difficulty (Easy, Medium, Hard, Extra) and are compared for SQL generation for different models. }
\label{fig:performance}
\end{figure}

Furthermore, Fig. \ref{fig:performance} represents the category-wise (Easy, Medium, Hard and Extra) SQL Generation results for different models.

\begin{table*}[htbp!]
\caption{Execution accuracy (EX) and exact set match accuracy (EM) on the holdout test set of Spider}
\label{tab:SQL_Accuracy}
\begin{center}
\begin{tabular}{lcc}
\toprule
\textbf{Model}                                                                 & \textbf{EX} & \textbf{EM} \\ \midrule

{DAIL-SQL + GPT-4 + Self-Consistency {\cite{b1}}}        & 86.6        & -          \\
{DAIL-SQL + GPT-4 {\cite{b1}}}        & 86.2        & -          \\
{DIN-SQL + GPT-4 {\cite{b2}}}        & 85.3        & 60          \\
{C3 + ChatGPT + Zero-Shot {\cite{b24}}}        & 82.3        & -          \\
{RESDSQL-3B + NatSQL (DB content used) {\cite{b29}}}        & 79.9        & 72.0          \\
{DIN-SQL +   CodeX davinci {\cite{b2}}}                                             & 78.2        & 57          \\
{Graphix-3B+PICARD (DB content used) {\cite{b31}} }           & 77.6        & \textbf{74}        \\
\textbf{End-to-end Text-to-SQL: Gpt-4o-mini (ours)}                                      &\textbf{92.44}       & 64.78       \\
\textbf{End-to-end Text-to-SQL: Gpt-3.5-turbo (ours)}                                           & \textbf{91.44}      & 60.45       \\
\bottomrule
\end{tabular}
\end{center}
\end{table*}

\subsection{Discussion}

\textbf{RQ1: What is the effect of db\_id prediction rules on db\_id prediction model?}

As shown in Table \ref{tab:cls model comparison}, the db\_id prediction model with NLQ has an initial accuracy of $86.38\%$ which increases by approximately $1.33\%$ when trained with NLQ and db\_id prediction rules. After analyzing the training data carefully we found that a few of the db\_id  are not separable e.g. college1, college2 etc. due to overlap of domain words. To make all the db\_id separable we have manually merged db\_id as explained in Section \ref{sect:train data prep}. Finally, when training the db\_id prediction model with merged db\_id,the db\_id prediction accuracy increased by approximately $3.85\%$ from the base model and by $1.54\%$ from the model trained with rules and NLQ without mergging. We can conclude that further rules for finetuning can make the db\_id prediction model scalable and can handle the non-separability and eliminate the need for manual merging of non-separable db\_id.

\textbf{RQ2: What is the impact of SQL correction on text-to-SQL?}

As part of the post-processing step, as detailed in Section \ref{sect:SQL-self-correction}, all SQL queries generated in stage 2 are further processed through the SQL correction module, which operates within a multi-agent framework. This module is designed to refine and correct the generated SQL queries, ensuring improved accuracy and adherence to database constraints.
The evaluation results, presented in Table \ref{tab: self-correction impacy}, demonstrate a notable improvement in both Execution Accuracy (EA) and Exact Match (EM), with increases of $18.44\%$ and $2.44\%$ for GPT-4o-Mini and $23.95\%$ and $2.66\%$ for GPT-3.5 respectively. These results clearly highlight the significant impact of our overall text-to-SQL generation framework, reinforcing the effectiveness of our multi-stage approach in enhancing query correctness and execution reliability.

\begin{table}[ht]
\caption{db\_id prediction Model accuracy comparison with base model}
\label{tab:cls model comparison}
\centering
\resizebox{\linewidth}{!}{
\begin{tabular}{lccc} 
\toprule 
\textbf{Eval Set} &\textbf{Base (RoBerta} & \textbf{Base With db\_id}  & \textbf{db\_id}  \\
&  \textbf{Based Encoder)} &\textbf{ Prediction Rules} & \textbf{Prediction after Merging} \\
\midrule
\textbf{Test}     & 86.38\%                      & 87.71\% (\textbf{+1.54\%})                          & 89.71\% (\textbf{+3.85\%}) \\ 
\textbf{Val}      & 85.71\%                      & 87.33(\textbf{+1.62})\%                           & 90.38\%(\textbf{+5.44\%}) \\
\bottomrule
\end{tabular}
}
\end{table}

\begin{table}[ht]
\caption{Impact of self-correction on execution accuracy (EX) and exact set match (EM)}
\label{tab: self-correction impacy}
\centering
\resizebox{\linewidth}{!}{ 
\begin{tabular}{l|cc|cc} 
\toprule
\textbf{}        & \multicolumn{2}{c}{\textbf{Before SQL Self-Correction}} & \multicolumn{2}{|c}{\textbf{After SQL Self-Correction}}  \\ 
\textbf{Model}   & \textbf{EX}                                  & \textbf{EM}        & \textbf{EX}                     & \textbf{EM}                   \\ \midrule
\textbf{GPT-4o-Mini}  & 74.00\%                                  & 62.74\%     & {92.44\%(\textbf{+18.44\%})}          & 64.78\%(\textbf{+2.04\%})         \\ 
\textbf{GPT-3.5} & 67.49\%                               & 57.79\%     & 91.44\%(\textbf{+23.95\%})          & 60.45\%(\textbf{+2.66\%})         \\ \bottomrule
\end{tabular}
}

\end{table}
\section{Conclusion and Future work}
We propose an end-to-end framework for text-to-SQL generation, addressing limitations in existing SOTA models where db\_id is manually provided for schema linking followed by a prompt-based mechanism for SQL generation. In contrast, our framework automatically predicts the db\_id, enabling a fully automated pipeline where only the NLQ is required as input to generate the corresponding SQL output. Extensive evaluations on the Spider dataset demonstrate that our approach significantly improves performance across all query complexity levels (easy, medium, hard, extra hard) with an average of approx $24\%$ and achieves results that are comparable or superior to SOTA LLM-based text-to-SQL models.

Our evaluation results indicate that the proposed approach demonstrates strong performance in schema linking; however, it exhibits a slight slowdown as the number of databases increases. This is primarily due to the growth in rule complexity, which in turn leads to an increase in prompt length. To mitigate this, our proposed framework employs a divide-and-conquer strategy for managing extended prompts. In future work, we plan to explore a Retrieval-Augmented Generation (RAG)-based approach, leveraging embedding-based similarity to dynamically retrieve the most relevant rules for a given NLQ  while maintaining a flexible similarity threshold.

\section*{Acknowledgment}
This work has received support from SUTD's Kickstart Initiative under grant number SKI 2021\_04\_06 and MOE under grant number MOE-T2EP20124-0014.

\vspace{12pt}

\end{document}